\documentclass[10pt,twocolumn,letterpaper]{article}

\usepackage{wacv}              

\usepackage{graphicx}
\usepackage{amsmath}
\usepackage{amssymb}
\usepackage{booktabs}
\usepackage{multirow}
\usepackage{array}
\usepackage[table,xcdraw]{xcolor}
\usepackage{colortbl}
\usepackage{chngcntr}

\usepackage[pagebackref,breaklinks,colorlinks]{hyperref}

\usepackage{algorithm}
\usepackage{algpseudocode}
\usepackage{listings}
\usepackage{xcolor}
\newcommand{\edit}[1]{\textcolor{red}{#1}}
\usepackage{cuted}
\usepackage{capt-of}

\usepackage[capitalize]{cleveref}
\crefname{section}{Sec.}{Secs.}
\Crefname{section}{Section}{Sections}
\Crefname{table}{Table}{Tables}
\crefname{table}{Tab.}{Tabs.}

\begin{document}

\title{ Rethinking Prompting Strategies for Multi-Label Recognition with Partial Annotations} %
\author{Samyak Rawlekar, Shubhang Bhatnagar, Narendra Ahuja\\
University of Illinois Urbana-Champaign\\
{\tt\small {samyakr2,sb56,n-ahuja}@illinois.org}
}
\maketitle

\begin{strip}
    \centering
    \includegraphics[width=1\linewidth]{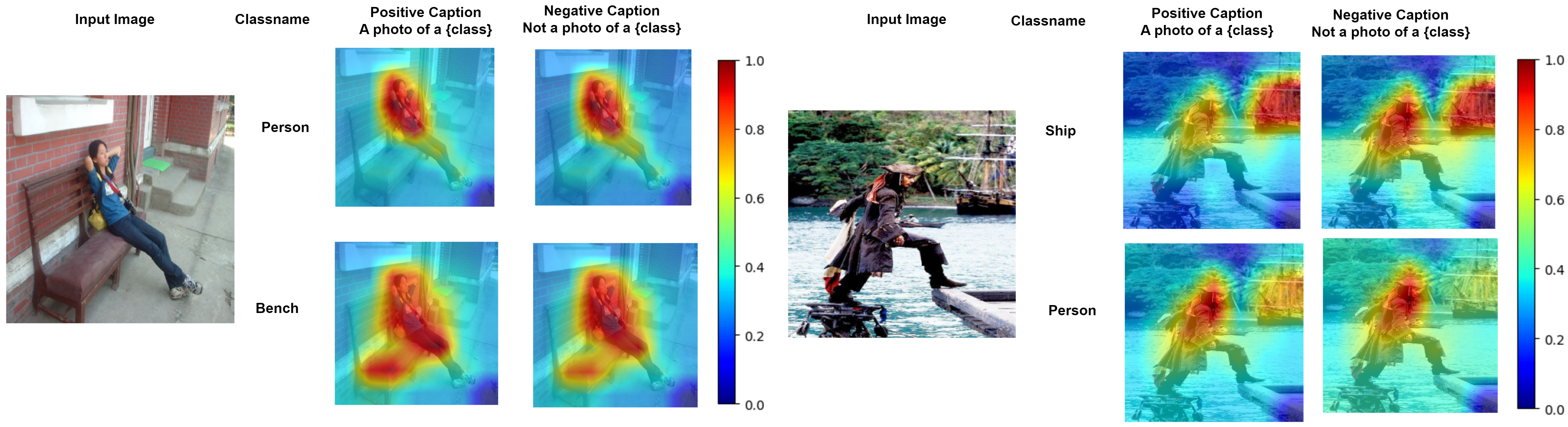}
    \captionof{figure}{ \textbf{Negative prompts with CLIP} Given an image, we visualize similarity of image features from CLIP \cite{clip} with both positive and negative captions for a class. Red regions highlight the activated areas, revealing that similar regions are activated by both positive and negative prompts. This indicates that CLIP associates the same features associated with presence of a class with both prompts, making us question the effectiveness of learning negative prompts using CLIP.}
    \label{fig:CLIP's similarity}
\end{strip}

\begin{abstract}
%
Vision-language models (VLMs) like CLIP have been adapted for Multi-Label Recognition (MLR) with partial annotations by leveraging prompt-learning, where positive and negative prompts are learned for each class to associate their embeddings with class presence or absence in the shared vision-text feature space. While this approach improves MLR performance by relying on VLM priors, we hypothesize that learning negative prompts may be suboptimal, as the datasets used to train VLMs lack image-caption pairs explicitly focusing on class absence.
To analyze the impact of positive and negative prompt learning on MLR, we introduce PositiveCoOp and NegativeCoOp, where only one prompt is learned with VLM guidance while the other is replaced by an embedding vector learned directly in the shared feature space without relying on the text encoder. Through empirical analysis, we observe that negative prompts degrade MLR performance, and learning only positive prompts, combined with learned negative embeddings (PositiveCoOp), outperforms dual prompt learning approaches. Moreover, we quantify the performance benefits that prompt-learning offers over a simple vision-features-only baseline, observing that the baseline displays strong performance comparable to dual prompt learning approach (DualCoOp), when the proportion of missing labels is low, while requiring half the training compute and 16 times fewer parameters. 

\end{abstract}



\begin{figure}
    \centering
    \includegraphics[width=1\linewidth]{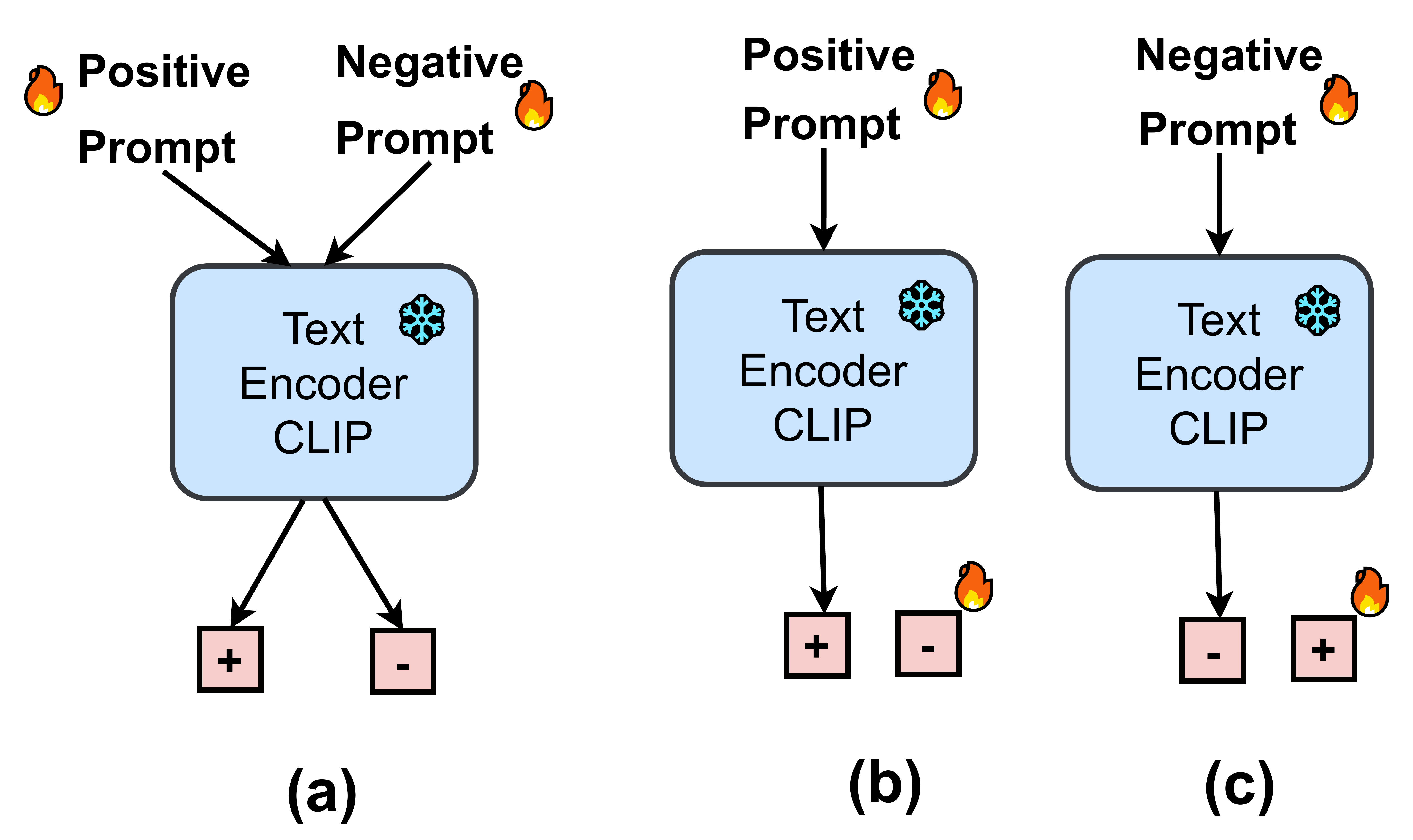}
    \caption{\textbf{Conceptual Comparison of MLR Approaches.} 
    In (a), we show the textual framework for existing VLM-based MLR approaches with partial annotations. They use CLIP's guidance to learn prompts for each class: a positive prompt associated with the presence of the class and a negative prompt associated with the absence of the class.
    To analyze the effect of the positive and negative guidance, we create two setups. In (b) we test the impact of positive guidance by removing the negative prompt and instead learn a negative embedding directly in feature space to detect class absence. The positive prompt, learned with CLIP, remains for detecting class presence. In (c), we test the impact of negative guidance by removing the positive prompt and instead learn a feature space embeddings to detect class presence. The negative prompt, learned with CLIP, is used to detect class absence. 
    }
    \label{fig:Difference from Existing approaches}
\end{figure}

\section{Introduction}
\label{sec:intro}

Multi-label recognition (MLR) is an important task in computer vision, being relevant to several real-world applications. Unlike single-label recognition, where an image is associated with just one label, MLR requires identifying multiple objects or concepts present in an image, making it significantly more challenging. Examples of MLR applications include medical diagnosis from chest X-rays \cite{huang2024radiology}, product detection in e-commerce images \cite{ecomm_product}, and food recognition for dietary monitoring systems \cite{nutrient2,nutrient3}.

The complexity of MLR arises from the combinatorial increase in the number of possible label subsets, which escalates the need for annotations. However, obtaining sufficient annotations is often infeasible, especially in the presence of class imbalance or rare classes \cite{rare}. This creates a pressing need to address MLR with partial annotations, which is crucial for improving the recognition performance of  real-world MLR systems.

Recent approaches \cite{dualcoop,MLR-CoOp,scpnet} have attempted to tackle this challenge by using large VLMs like CLIP \cite{clip}. Such approaches make use of CLIP’s vision and text encoders that have been trained on much larger and more diverse data than previous approaches that used ImageNet based initializations. Many of these strategies rely on parameter-efficient prompt-learning strategies while keeping the weights of the VLM frozen when fine-tuning on smaller datasets. Such approaches have demonstrated significant performance improvements while being efficient to train.
Prompt learning particularly helps take advantage of the latent knowledge in the text encoder to learn prompts whose embeddings in the shared vision-language embedding space correspond to the presence of a class.
DualCoOp \cite{dualcoop} extended prompt learning to MLR, learning a positive and negative prompts for each class. Several MLR \cite{dualcoop++,MLR-CoOp} approaches borrow a similar dual prompt learning framework. Similarity of image embedding with the text embedding of the positive prompt indicates the presence of a class, while similarity of image embedding with the text embedding of the negative prompt indicates its absence. Learning such a dual prompt framework has been shown to benefit MLR when partial annotations are available.

However, a closer analysis of VLMs reveals that they have been trained on image-caption datasets (Sec. \ref{Sec:Analysis}) where captions correspond to the presence of objects rather than their absence. This also reflects in the text embeddings produced by CLIP for simple handcrafted negative prompts being more highly correlated with regions with the object present like positive prompts as seen in Fig \ref{fig:CLIP's similarity}.
These make us question the utility of using \textbf{guidance from VLMs to learn negative prompts} in such approaches. We conduct a thorough empirical study to evaluate the contribution of VLM guidance in learning negative prompts in such partial annotation MLR settings. Specifically, we use two hybrid setup: (a) Positive CoOp and (b) Negative CoOp. In Postive CoOp (Negative CoOp), (1) Like existing methods, we learn a positive (negative) text prompt, whose embedding obtained using the VLM text encoder is associated with image features indicating presence (absence) of the class and (2)  Learn a negative (positive) embedding that corresponds to image features indicating absence (presence) of the class in the shared feature space. Note that learning such a \textit{negative (positive) embedding makes no use of the text encoder} information/guidance, while also allowing us to still keep the rest of the setup from previous dual prompt based MLR works intact for an accurate comparison. 

We also additionally compare the performance of these methods with a simple non-prompting based baseline that only uses the vision encoder of a VLM, and is trained using state-of-the art MLR loss and optimization techniques. Previous works do not provide such a simple baseline for comparison.

We evaluate our one-layer linear projector baseline, Positive CoOp and the Negative CoOp setup on two standard multi-label recognition (MLR) benchmark datasets: COCO \cite{coco} and VOC2007 \cite{pascal-voc}. We find that Positive CoOp outperforms DualCoOp, while Negative CoOp shows even lesser performance than the baseline that uses no text information. We also find that the baseline shows strong performance, especially comparable to that of DualCoOp and positive CoOp when the proportion of available labels is high (60\%-90\%). The baseline also requires approximately 16 times fewer training parameters than DualCoOp, half training hours on COCO, due to it only relying on frozen visual features and not backpropagating through the text encoder.

To summarize, our \textbf{contributions} are:
\begin{itemize}
     \item  We thoroughly analyze the impact of negative prompting on MLR in partial annotations, demonstrating for the first time that their use adversely affects model performance.
     \item We show that only using positive prompts and replacing negative prompt learning with learning negative embeddings without guidance from VLM text encoder outperforms the dual-prompt learning for MLR in partial annotations.
    \item We propose a simple baseline that only uses the vision encoder of a VLM, and use it to accurately quantify the performance gains of positive and negative prompting. We show that such a baseline shows strong performance while also requiring 16 times fewer training parameters and half the GPU hours than the prompt learning methods while training on COCO, and 15 times fewer parameters and half the GPU hours while training on VOC2007.
\end{itemize}

\section{Related Works}
\label{sec:related works}
\noindent\textbf{Multi-Label Recognition with Partial Annotations:}
Our work aims to recognize multiple objects within an image, similar to many previous efforts in the field\cite{SST,cole2021multi,rnn2, rnn1, rnn4}. Due to the difficulty in annotations, we are particularly interested in MLR with partial annotations, where some classes may not be labeled in each image. Early approaches  tackled this problem by ignoring the missing labels and training disjoint binary classifiers for each object on the known label set\cite{boostexter,liu2015optimality,misra2016seeing}. However, the performance of these methods during inference was significantly hindered by the partial or missing annotations. To overcome this, subsequent work proposed the idea of replacing the missing annotations with pseudo labels \cite{partial-label-learning1,partial-label-learning2,partial-label-learning3}. They use pretrained models or train a new model with modified loss to classify the missing labels into the known set. Most recent line of works explictly transfer information of labels from one image to another by using either label dependencies \cite{tsoumakas2006multi, gcnmulti} or by blending features of known labels from one image to the unknown labels in another \cite{SARB}. While these approaches have made notable advances, they typically require large MLR datasets or tailored loss functions, and they still struggle with very low percentages (10\%-30\%) of available labels. On the other hand, VLM-based approaches demonstrate that VLM priors help them achieve higher performance even with only 10\%-30\% of the available labels.  \\

\noindent\textbf{Vision-Language Models for MLR with Partial Annotations}: 
Over the past two years, MLR has increasingly focused on adopting VLMs such as CLIP \cite{clip}. VLMs learn representations by aligning hundreds of millions of image-text pairs, enabling them to adapt to various downstream tasks such as classification \cite{adapter1, adapter2, adapter3, adapter4}, retrieval \cite{ret, ret2}, and segmentation \cite{seg2}. SCPNet \cite{scpnet} leverages class name similarities derived from CLIP’s embedding space and augments the training with a self-supervised contrastive loss. Furthermore, inspired by large language models (LLMs), recent approaches use prompting to adapt VLMs for MLR. Specifically, \cite{dualcoop} learns two prompts per class—one for presence and one for absence—while \cite{dualcoop++} extends this by introducing a third prompt called the evidence prompt. These prompt embeddings are used to detect class presence or absence in local image regions, which are then aggregated for predictions. Despite the significant performance gains and reduced parameter requirements, the performance is still suboptimal due the use of negative prompt. We show that, the PositiveCoOp setup which learns only the positive prompt with guidance from CLIP and learn to associate the absence of class with the embeddings learned without CLIP's guidance can outperform dual prompt based approaches while requiring even fewer parameters and less GPU training hours.

\section{Approach}

\begin{figure}
    \centering
    \includegraphics[width=1\linewidth]{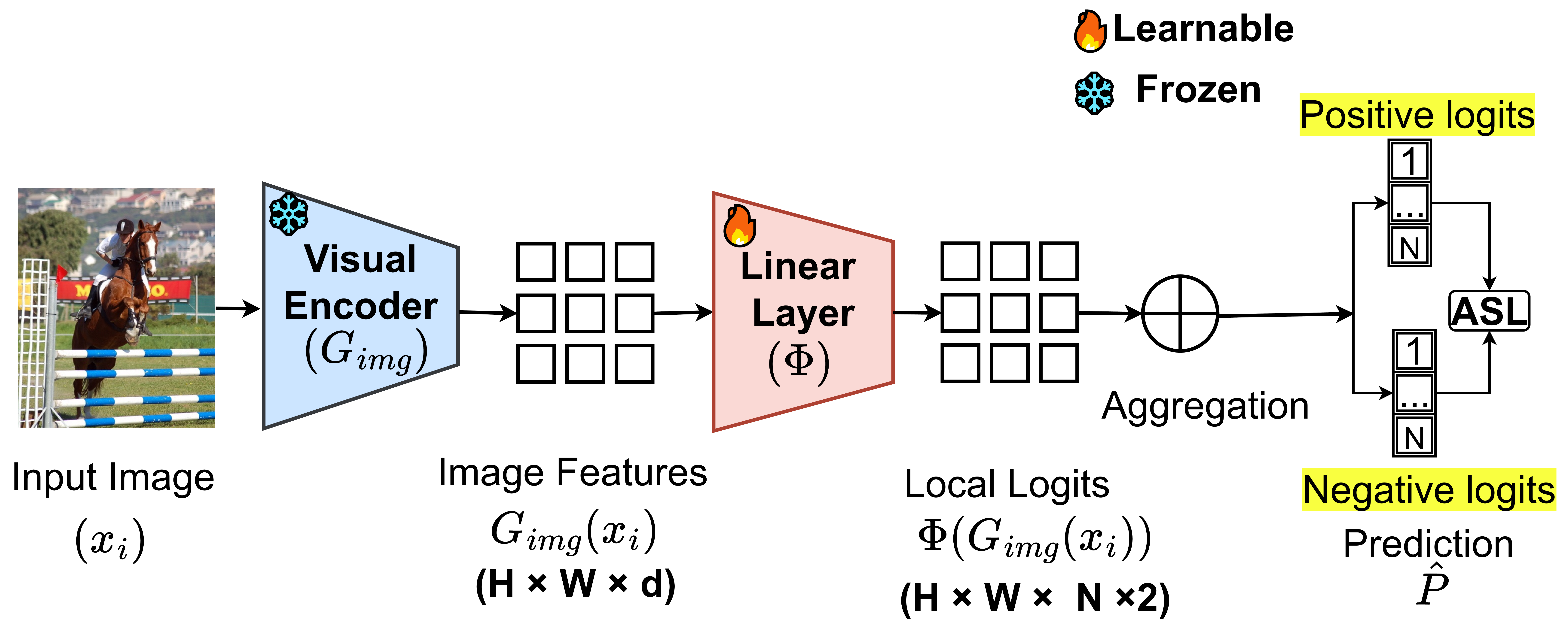}
    \caption{\textbf{Baseline Framework.} To quantify the impact of prompting based approaches in MLR with partial annotations, we setup up a baseline (sec \ref{subsec:baseline}) that uses only visual information. Given an image $\mathbf{x}_{i}$, with multiple objects, we first extract its features ($G_{\text{img}}(\mathbf{x}_i)$) using the frozen visual encoder of CLIP \cite{clip}. These features are then passed through a linear projector layer ($\Phi$) that projects the d-dimensional features at location $(h,w)$ to two local logits per class for all $N$ classes, one logit indicating the presence of the class and another its absence. The local logits are aggregated across all spatial regions to produce the final positive and negative logits. We train the linear projector layer of the baseline using the widely used asymmetric loss \cite{asl}.}
    \label{fig:baseline}
\end{figure}

\begin{figure*}
    \centering
    \includegraphics[width=0.8\linewidth]{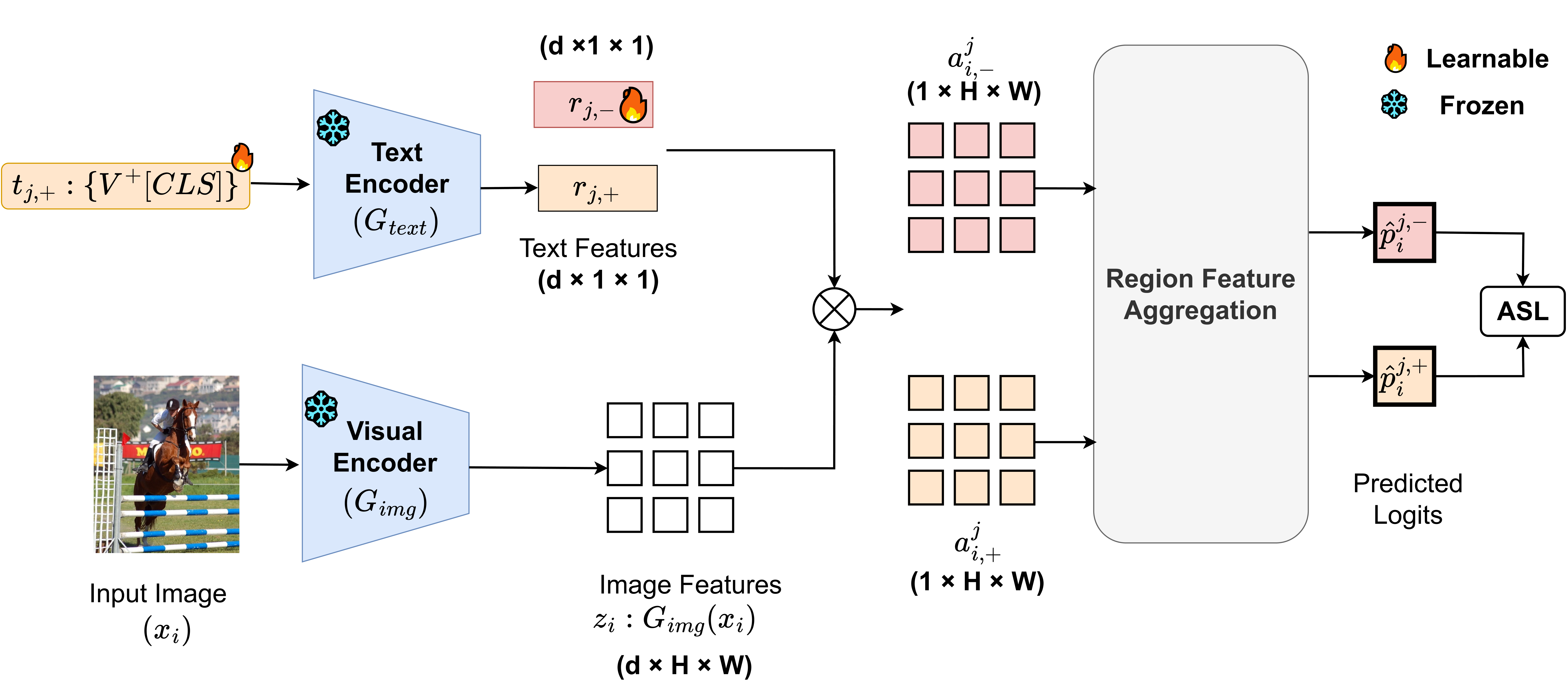}
    \caption{\textbf{PositiveCoOp and NegativeCoOp Overview.}
   This figure illustrates the PositiveCoOp framework, with NegativeCoOp being its mirror image. 
    VLM based MLR approaches like DualCoOp \cite{dualcoop} propose to learn both positive and negative prompts using CLIP's guidance: one for class presence and one for class absence.
     In PositiveCoOp (NegativeCoOp), for a given class j only the positive (negative) prompt \(\mathbf{t}_{j,+}\) (\(\mathbf{t}_{j,-}\)) is learned through CLIP, while the negative (positive) prompt is replaced by a learned text embedding \(\mathbf{r}_{j,-}\) (\(\mathbf{r}_{j,+}\)) in the feature space, independent of CLIP's text encoder. 
    For both PositiveCoOp and NegativeCoOp, we obtain the final predictions $\mathbf{\hat{p}_i^{j,+}}$ and $\mathbf{\hat{p}_i^{j,-}}$ by calculating the cosine similarity of the image features with the embedding of the positive text prompt $\mathbf{r}_{j,+}$ and learned text embedding $\mathbf{r}_{j,-}$ and then aggregating this using the the class specific feature aggregation strategy following \cite{dualcoop}  , described in detail Sec. \ref{subsec:baseline}. Only the text embeddings and the prompts are trained using the widely used Asymmetric Loss \cite{asl}
    }
    \label{fig:Overview}
\end{figure*}

 Our primary objective is to validate the hypothesis presented in Sec \ref{sec:intro}, which argues that CLIP's guidance to learn a negative prompt is not optimal and, in fact it reduces performance. 

 To investigate this, we breakdown and analyze the effect of various components of the prompting strategies in SOTA VLM-based MLR methods which operate in partial annotation settings. Specifically, we focus on prompting strategies of DualCoOp, that leverages CLIP to learn a positive and a negative prompt. The positive prompt is associated with the presence of the class and the negative prompt is associated with the absence of the class. 
 %
 To quantify the impact of these individual learned prompts, we first establish a baseline ($B$) (sec \ref{subsec:baseline}), that use only the visual information, completely omitting any form of text-based input. This baseline design uses the same visual encoders as existing MLR with partial annotations setups to ensure a fair and a direct comparison with approaches that use prompts. To measure the individual impact of positive and negative prompts, we introduce two setups (sec. \ref{subsec:Positive CoOp and Negative CoOp}): (1) PositiveCoOp, which uses CLIP to learn only positive prompts, and (2) NegativeCoOp, which learns only negative prompts with CLIP's guidance. 
 
 In PositiveCoOp (NegativeCoOp), instead of using CLIP's guidance to learn a negative (positive) prompt like DualCoOp and other related methods\cite{dualcoop++} we learn a negative (positive) text embedding without using text encoder guidance that corresponds to absence (presence) of the class. The conceptual difference of PositiveCoOp and NegativeCoOp with existing prompting based methods is shown in Fig. \ref{fig:Difference from Existing approaches}. 
 
 Lastly, following \cite{dualcoop,MLR-CoOp,dualcoop++,asl}, we train all three setups (Baseline, PositiveCoOp and NegativeCoOp) using the widely used asymmetric loss function \cite{asl}. An overview of the baseline is presented in Fig. \ref{fig:baseline}, while  Fig.\ref{fig:Overview} presents the overview of PositiveCoOp and NegativeCoOp.
 
\textbf{Formulation: MLR with partial annotation.}  Given dataset \(\mathcal{D}\) consisting of images \(\{\mathbf{x}_{i}\}_{i=1}^{|\mathcal{D}|}\), and \(N\) classes \(\{C_j\}_{j=1}^N\). Each image (\(\mathbf{x}_i\)) can be associated with one or more of these \(N\) classes. The MLR task is to identify the subset of classes \(\mathcal{C}_i \subseteq \{C_1, C_2, \ldots, C_N\}\) corresponding to each image \(\mathbf{x}_i\) by learning a function \(f: \mathbf{x_{i}} \rightarrow \{1,-1\}^N\), such that the output vector \(f(\mathbf{x}_i)\) maps to (\(1\)) if the class is present, (\(-1\)) if the class is absent. In the context of MLR with partial annotations, we face the additional challenge that only a subset of the classes associated with each image is annotated during training. Despite these incomplete annotations, our objective during inference remains to accurately identify all classes present in the image.

In all three setups, the VLM (CLIP), denoted by $G$, is frozen. $G$ consists of two encoders: one for encoding images (\(G_{\text{img}}\)) and another for encoding text (\(G_{\text{text}}\)).

\subsection{Baseline} \label{subsec:baseline}

We propose a Baseline ($B$), to quantify the impact of different prompting strategies used in VLM based MLR methods that operate with partial annotations. This baseline only uses the visual information from the large VLMs and is trained using the widely use Asymmetric Loss\cite{asl}.

To establish this baseline, we use only the visual encoder (\(G_{\text{img}}\)) of CLIP for feature extraction, and use these features for MLR without using the text encoder ($G_{\text{text}}$). Specifically, our visual encoder of the baseline ($B$) follows standard CLIP visual encoder setup of VLM-based MLR methods that operate in partial annotations setting, which involves removing the final pooling layer and obtaining spatial features.  The final pooling layer is removed to preserve class-specific information across spatial regions, which could otherwise be destroyed by pooling, as pooled features are often dominated by features of a single class, which is not suitable for MLR. 

Using $G_{img}$, we obtain the feature map ($\mathbf{z}_i$) for an image ($\mathbf{x}_i$), where $\mathbf{z}_i = G_{\text{img}}(\mathbf{x}_i) \in \mathbb{R}^{H \times W \times d}$ with height $H$, width $W$, and feature dimension $d$. For each $d$-dimensional feature at spatial location $(h, w)$ in $\mathbf{z}$, denoted as $\mathbf{z}(h,w,:)$, we train a linear projector layer ($\Phi$), that projects the d-dimensions at each location $(h,w)$ to $N\times 2$ dimension at that location. This $N\times 2$ corresponds to two local logits per class for each of the $N$ classes: one indicating the presence of the class and the other indicating its absence. 

The local logits at a location $(h,w)$ is given by $\mathbf{a}_i (h,w) = \Phi(\mathbf{z}_i(h,w,:)) \in \mathbb{R}^{N \times 2}$.  
The overall local logits at all locations is given by 
$\mathbf{a}_i \in  \mathbb{R}^{H \times W \times (N \times 2)}$
\[\text{Positive Local Logits: } \mathbf{a}_i^+ = \mathbf{a}_i[:,:,:,0] \in \mathbb{R}^{H \times W \times N}\] 
\[ \text{Negative Local Logits: } \mathbf{a}_i^- = \mathbf{a}_i[:,:,:,1] \in \mathbb{R}^{H \times W \times N}\] 

Following \cite{dualcoop}, we obtain the softmax map from $\mathbf{a}_i^+$ and $\mathbf{a}_i^-$, which assign weights to different regions such that they pay more attention to regions that contain classes for $\mathbf{a}_i^+$ and regions that do not contain classes for $\mathbf{a}_i^-$. The softmax map for each class $n$ at location ($h,w$) is given by  
\[
\mathbf{A}^+_n[h,w] = \frac{\exp(\mathbf{a^+}_i[h,w])}{\sum_{h'=1}^{H}\sum_{w'=1}^{W} \exp(\mathbf{a^+}_i[h',w'])}.
\]

\[
\mathbf{A}^-_n[h,w] = \frac{\exp(\mathbf{a^-}_i[h,w])}{\sum_{h'=1}^{H}\sum_{w'=1}^{W} \exp(\mathbf{a^-}_i[h',w'])}.
\]

The prediction (\(\mathbf{\hat{p}_i} = [\mathbf{\hat{p}_i^+}, \mathbf{\hat{p}_i^-}]\) ) is obtained by the dot product of the softmax map ($\mathbf{A}^+_n, \mathbf{A}^-_n$) with the local logits (\(\mathbf{a}^+_i, \mathbf{a}^-_i \)) and aggregating over the spatial regions:

\[
\mathbf{\hat{p}}^+_i[n] = \sum_{h=1}^{H} \sum_{w=1}^{W} \mathbf{A}^+_n[h,w] \cdot \mathbf{a}^+_i[h,w].
\]
\[
\mathbf{\hat{p}}^-_i[n] = \sum_{h=1}^{H} \sum_{w=1}^{W} \mathbf{A}^-_n[h,w] \cdot \mathbf{a}^-_i[h,w].
\]
where, $n \in \{1, \ldots, N\}$

\subsection{PositiveCoOp and NegativeCoOp}\label{subsec:Positive CoOp and Negative CoOp} 

With the baseline established using only visual information, we now explore how incorporating textual information affects performance. DualCoOp leverages textual information from CLIP to enhance visual understanding by learning a positive and a negative prompt for each class $j$, denoted as (${ \mathbf{t_{j,+}}, \mathbf{t_{j,-}} }$), by backpropagating them through the text encoder ($G_\text{text}$). PositiveCoop and NegativeCoOp are ablations of DualCoOp designed to isolate the effect of positive and negative guidance respectively.

\textbf{PositiveCoOp. } Following prompt-learning approaches \cite{dualcoop, dualcoop++, MLR-CoOp}, for each class $j$, we initialize a positive prompt \(\mathbf{t}_{j,+}\) using the template $[V^+]\{\text{classname}\}$, where $V$ is the learnable word embedding vector that maximizes the cosine similarity between image and text features to improve recognition. The positive prompt \(\mathbf{t}_{j,+}\) is passed through the frozen text encoder ($G_{\text{text}}$) to produce an embedding $\mathbf{r}_{j,+} \in \mathbb{R}^d$. The cosine similarity between the image features ($\mathbf{z}_i$) and the embeddings ($\mathbf{r}_{j,+}$) indicates the presence of class $j$ in the image. In contrast to existing approaches, we do not use negative prompt. Instead, we learn an embedding ($\mathbf{r}_{j,-} \in \mathbb{R}^d$ ) in feature space, trained to provide negative evidence for that class without any guidance from CLIP, as shown in Fig. \ref{fig:Overview}. The cosine similarity of image features ($\mathbf{z}_i$) with $\mathbf{r}_{j,-}$ indicates the absence of class $j$.

\textbf{NegativeCoOp.} In contrast to PositiveCoOp, this setup evaluates how CLIP guides the learning of a negative prompt, while we remove the positive prompt and learn it directly in the embedding space. Specifically, following \cite{dualcoop, dualcoop++, MLR-CoOp}, for each class $j$, we initialize a negative prompt (\(\mathbf{t}_{j,-}\)) using the template $[V^-]\{\text{classname}\}$ which is passed through the frozen text encoder ($G_{\text{text}}$) to produce an embedding $\mathbf{r}_{j,-} \in \mathbb{R}^d$. 
We do not use the positive prompt to detect presence of a class, and instead learn an embedding ($\mathbf{r}_{j,+} \in \mathbb{R}^d$ ) in feature space, trained to provide positive evidence for class $j$, without any guidance from CLIP.  The cosine similarity of image features ($\mathbf{z}_i$) with $\mathbf{r}_{j,+}$ indicates the presence of class $j$, while the similarity between $\mathbf{r}_{j,-}$ and $\mathbf{z}_i$ indicates the absence of the class in the image.

To obtain the predictions ($\mathbf{\hat{p}} \in \mathbb{R}^{(N \times 2)}$) for each of PositiveCoOp and NegativeCoOp, we follow the procedure described in sec.\ref{subsec:baseline} and in \cite{dualcoop}. This includes class specific region feature aggregation, which takes in the input the image features ($\mathbf{z}_i$) and the embeddings ($\mathbf{r}_{j,+}$ and $\mathbf{r}_{j,-}$), computes the dot product between image and text features to obtain positive map, $\mathbf{a}_{i,+}^j = $($\mathbf{z}_i \cdot \mathbf{r}_{j,+}$) and 
negative map, $\mathbf{a}_{i,-}^j = (\mathbf{z}_i \cdot \mathbf{r}_{j,-})$, followed by product with softmax map to assign more focus to the regions that contain class $j$ for $\mathbf{a}_{i,+}^j$ and to regions that do not contain class $j$ for $\mathbf{a}_{i,-}^j$  (described in sec. \ref{subsec:baseline})
and aggregation along spatial dimension to get $\mathbf{\hat{p}^{j,+}_i} \in \mathbf{R}$ and $\mathbf{\hat{p}^{j,-}_i} \in \mathbf{R}$, together to obtain the positive and negative logits for class $j$, $\mathbf{\hat{p}^j_i}$: [$\mathbf{\hat{p}^+_i}$, $\mathbf{\hat{p}^-_i}$] and training with the widely used ASL loss.

\subsection{Training}
\label{subsec:training}

Our works comprises of three setups:  Baseline, PositiveCoOp and NegativeCoOp. The visual ($G_{\text{img}}$) and textual encoder ($G_{\text{text}}$)  is frozen for all these setups.

For Baseline, we only train the linear projector layer ($\Phi$). For PositiveCoOp, we train the positive prompt and the embedding $\mathbf{r}_{j,-}$ in the embedding space. For NegativeCoOp, we learn negative prompt and the embedding $\mathbf{r}_{j,+}$ in the embedding space. We train all setups using widely used Asymmetric Loss (ASL) \cite{asl}.


ASL \cite{asl} is designed to address the inherent imbalance in multi-label recognition (MLR) caused by the significantly higher number of negative examples compared to positive ones in the training images. Following focal loss \cite{focal}, ASL down-weights the loss contribution from negative examples. However, instead of using a single focusing parameter ($\gamma$) as in focal loss, ASL uses two focusing parameters ($\gamma_{+}$ and $\gamma_{-}$ ).
\begin{align}
    \mathcal{L}_{ASL}(\hat{y}_{i}^{j}) = \begin{cases}
 &  \left(1-\hat{y}_{i}^{j}\right)^{\gamma_{+}} \log \left(\hat{y}_{i}^{j} \right) \text{ when } y_{i}^{j} =1 \\
     &  \left(\hat{y}_{i, \delta}^{j}\right)^{\gamma_{-}}  \log \left(1-\hat{y}_{i, \delta}^{j}\right) \text{ when } y_{i}^{j}=0
\end{cases} 
\vspace{-6pt}
\label{eq:instance_RASL}
\end{align}
where $\hat{y}_{i}^{j}$ represents the corresponding prediction associated with label $y_{i}^{j}$; $\hat{y}_{i, \delta}^{j} = \max(\hat{y} - \delta, 0)$, with $\delta$ representing the shifting parameter defined in ASL.

For all three setup, the class-specific prediction vector $\mathbf{\hat{p}} \in \mathbb{R}^{(N \times 2)} $ contains two entries per class for all $N$ classes, where positive logit vector ($\mathbf{\hat{p}[:,0]}$) corresponds to the presence of the class and the negative logits vector ($\mathbf{\hat{p}[:,1]}$) corresponds to its absence. The positive logit vector ($\mathbf{\hat{p}[:,0]}$) corresponds to $\hat{y}_{i}^{j}$, and the negative logit vector ($\mathbf{\hat{p}[:,1]}$) corresponds to ($1 - \hat{y}_{i}^{j}$) in Equation \ref{eq:instance_RASL}.

\noindent The total loss over the dataset $|\mathcal{D} |$ is given by:
\[
\mathcal{L}_{total} = \sum_{i=1}^{|\mathcal{D}|} \sum_{j=1}^{N} \mathcal{L}_{ASL}(\hat{y}_{i}^{j})
\]

\begin{table*}[ht]
    \centering
    
    \begin{tabular}{l|l|c|ccccccccc|c}
        \hline
        Dataset & Methods & \#Params & 10\% & 20\% & 30\% & 40\% & 50\% & 60\% & 70\% & 80\% & 90\% & Avg. \\ 
        \hline
        \multirow{11}{*}{\parbox{2.2cm}{COCO \cite{coco} \\ \textbf{(ResNet101)}}}
        & \cellcolor{lightgray!45}
        SSGRL \cite{SSGRL} & \cellcolor{lightgray!45}64.7M & \cellcolor{lightgray!45}62.5 & \cellcolor{lightgray!45}70.5 & \cellcolor{lightgray!45}73.2 & \cellcolor{lightgray!45}74.5 & \cellcolor{lightgray!45}76.3 & \cellcolor{lightgray!45}76.5 & \cellcolor{lightgray!45}77.1 & \cellcolor{lightgray!45}77.9 & \cellcolor{lightgray!45}78.4 & \cellcolor{lightgray!45}74.1 \\ 
        & \cellcolor{lightgray!45}
        GCN-ML \cite{gcnmulti} & \cellcolor{lightgray!45}44.9M & \cellcolor{lightgray!45}63.8 & \cellcolor{lightgray!45}70.9 & \cellcolor{lightgray!45}72.8 & \cellcolor{lightgray!45}74.0 & \cellcolor{lightgray!45}76.7 & \cellcolor{lightgray!45}77.1 & \cellcolor{lightgray!45}77.3 & \cellcolor{lightgray!45}78.3 & \cellcolor{lightgray!45}78.6 & \cellcolor{lightgray!45}74.4 \\
        & \cellcolor{lightgray!45}KGGR \cite{kggr} & \cellcolor{lightgray!45}$\geq$ 25M & \cellcolor{lightgray!45}66.6 & \cellcolor{lightgray!45}71.4 & \cellcolor{lightgray!45}73.8 & \cellcolor{lightgray!45}76.7 & \cellcolor{lightgray!45}77.5 & \cellcolor{lightgray!45}77.9 & \cellcolor{lightgray!45}78.4 & \cellcolor{lightgray!45}78.7 & \cellcolor{lightgray!45}79.1 & \cellcolor{lightgray!45}75.6 \\
        & \cellcolor{lightgray!45}CL \cite{partial-label-learning3} & \cellcolor{lightgray!45}$\geq$ 38M & \cellcolor{lightgray!45}26.7 & \cellcolor{lightgray!45}31.8 & \cellcolor{lightgray!45}51.5 & \cellcolor{lightgray!45}65.4 & \cellcolor{lightgray!45}70.0 & \cellcolor{lightgray!45}71.9 & \cellcolor{lightgray!45}74.0 & \cellcolor{lightgray!45}77.4 & \cellcolor{lightgray!45}78.0 & \cellcolor{lightgray!45}60.7 \\
        & \cellcolor{lightgray!45}Partial BCE \cite{partial-label-learning3} & \cellcolor{lightgray!45}$\geq$ 38M & \cellcolor{lightgray!45}61.6 & \cellcolor{lightgray!45}70.5 & \cellcolor{lightgray!45}74.1 & \cellcolor{lightgray!45}76.3 & \cellcolor{lightgray!45}77.2 & \cellcolor{lightgray!45}77.7 & \cellcolor{lightgray!45}78.2 & \cellcolor{lightgray!45}78.4 & \cellcolor{lightgray!45}78.5 & \cellcolor{lightgray!45}74.7 \\
        & \cellcolor{lightgray!45}SST \cite{SST} & \cellcolor{lightgray!45}33.5M & \cellcolor{lightgray!45}68.1 & \cellcolor{lightgray!45}73.5 & \cellcolor{lightgray!45}75.9 & \cellcolor{lightgray!45}77.3 & \cellcolor{lightgray!45}78.1 & \cellcolor{lightgray!45}78.9 & \cellcolor{lightgray!45}79.2 & \cellcolor{lightgray!45}79.6 & \cellcolor{lightgray!45}79.9 & \cellcolor{lightgray!45}76.7 \\
        & \cellcolor{lightgray!45}SARB \cite{SARB} & \cellcolor{lightgray!45}29.6M & \cellcolor{lightgray!45}71.2 & \cellcolor{lightgray!45}75.0 & \cellcolor{lightgray!45}77.1 & \cellcolor{lightgray!45}78.3 & \cellcolor{lightgray!45}78.9 & \cellcolor{lightgray!45}79.6 & \cellcolor{lightgray!45}79.8 & \cellcolor{lightgray!45}80.5 & \cellcolor{lightgray!45}80.5 & \cellcolor{lightgray!45}77.9 \\
        & SST* & 33.5M & 69.1 & 78.5 & 79.3 & 79.9 & 80.1 & 80.5 & 81.1 & 80.7 & 80.7 & 78.9 \\
        & SARB* & 29.6M & 75.5 & 78.5 & 79.0 & 79.5 & 80.4 & 80.2 & 80.8 & 80.6 & 80.8 & 79.4 \\
        & DualCoOp \cite{dualcoop} & 1.3M & 78.7 & 80.9 & 81.7 & 82.0 & 82.5 & 82.7 & 82.8 & 83.0 & 83.1 & 81.9 \\
        & SCPNet \cite{scpnet} &3.4M & \textbf{80.3} & \textbf{82.2} & 82.8 & 83.4 & \textbf{83.8} & 83.9 & 84.0 & 84.1 & 84.2 & \textbf{83.2} \\
        \cline{2-13}
        & Baseline  & 80k &  78.9&  80.6&  81.3&  81.9 &  82.7&  82.8&  82.9&  83.2 &  83.5& 82.0 \\
        & Negative CoOp  & 730k &  77.8&  80.3&  81.0&  81.9 &  82.2&  82.4&  82.7&  82.8 &  82.9& 81.6 \\
        & Positive CoOp  & 730k &  79.8&  82.1&  \textbf{83.0}&  \textbf{83.5} &  83.7&  \textbf{83.9}&  \textbf{84.0}&  \textbf{84.2} &  \textbf{84.4}& \textbf{83.2} \\

        \midrule
        \multirow{10}{*}{\parbox{2.2cm}{VOC2007\cite{pascal-voc} \\ \textbf{(ResNet101)} }}
        & \cellcolor{lightgray!45} 
        SSGRL \cite{SSGRL} & \cellcolor{lightgray!45}66.6M & \cellcolor{lightgray!45}77.7 & \cellcolor{lightgray!45}87.6 & \cellcolor{lightgray!45}89.9 & \cellcolor{lightgray!45}90.7 & \cellcolor{lightgray!45}91.4 & \cellcolor{lightgray!45}91.8 & \cellcolor{lightgray!45}91.9 & \cellcolor{lightgray!45}92.2 & \cellcolor{lightgray!45}92.2 & \cellcolor{lightgray!45}89.5 \\ 
        
        & \cellcolor{lightgray!45}GCN-ML \cite{gcnmulti} & \cellcolor{lightgray!45}44.9M & \cellcolor{lightgray!45}74.5 & \cellcolor{lightgray!45}87.4 & \cellcolor{lightgray!45}89.7 & \cellcolor{lightgray!45}90.7 & \cellcolor{lightgray!45}91.0 & \cellcolor{lightgray!45}91.3 & \cellcolor{lightgray!45}91.5 & \cellcolor{lightgray!45}91.8 & \cellcolor{lightgray!45}92.0 & \cellcolor{lightgray!45}88.9 \\
        & \cellcolor{lightgray!45}KGGR \cite{kggr} & \cellcolor{lightgray!45}$\geq$ 25M & \cellcolor{lightgray!45}81.3 & \cellcolor{lightgray!45}88.1 & \cellcolor{lightgray!45}89.9 & \cellcolor{lightgray!45}90.4 & \cellcolor{lightgray!45}91.2 & \cellcolor{lightgray!45}91.3 & \cellcolor{lightgray!45}91.5 & \cellcolor{lightgray!45}91.6 & \cellcolor{lightgray!45}91.8 & \cellcolor{lightgray!45}89.7 \\
        & \cellcolor{lightgray!45}CL \cite{partial-label-learning3} & \cellcolor{lightgray!45}$\geq$ 38M & \cellcolor{lightgray!45}44.7 & \cellcolor{lightgray!45}76.8 & \cellcolor{lightgray!45}88.6 & \cellcolor{lightgray!45}90.2 & \cellcolor{lightgray!45}90.7 & \cellcolor{lightgray!45}91.1 & \cellcolor{lightgray!45}91.6 & \cellcolor{lightgray!45}91.7 & \cellcolor{lightgray!45}91.9 & \cellcolor{lightgray!45}84.1 \\
        & \cellcolor{lightgray!45}Partial BCE \cite{partial-label-learning3} & \cellcolor{lightgray!45}$\geq$ 38M & \cellcolor{lightgray!45}80.7 & \cellcolor{lightgray!45}88.4 & \cellcolor{lightgray!45}89.9 & \cellcolor{lightgray!45}90.7 & \cellcolor{lightgray!45}91.2 & \cellcolor{lightgray!45}91.8 & \cellcolor{lightgray!45}92.3 & \cellcolor{lightgray!45}92.4 & \cellcolor{lightgray!45}92.5 & \cellcolor{lightgray!45}90.0 \\
        & \cellcolor{lightgray!45}SST \cite{SST} & \cellcolor{lightgray!45}32.4M & \cellcolor{lightgray!45}81.5 & \cellcolor{lightgray!45}89.0 & \cellcolor{lightgray!45}90.3 & \cellcolor{lightgray!45}91.0 & \cellcolor{lightgray!45}91.6 & \cellcolor{lightgray!45}92.0 & \cellcolor{lightgray!45}92.5 & \cellcolor{lightgray!45}92.6 & \cellcolor{lightgray!45}92.7 & \cellcolor{lightgray!45}90.4 \\
        &\cellcolor{lightgray!45}SARB \cite{SARB} & \cellcolor{lightgray!45}29.6M & \cellcolor{lightgray!45}83.5 & \cellcolor{lightgray!45}88.6 & \cellcolor{lightgray!45}90.7 & \cellcolor{lightgray!45}91.4 & \cellcolor{lightgray!45}91.9 & \cellcolor{lightgray!45}92.2 & \cellcolor{lightgray!45}92.6 & \cellcolor{lightgray!45}92.8 & \cellcolor{lightgray!45}92.9 & \cellcolor{lightgray!45}90.7 \\
        & DualCoOp \cite{dualcoop} & 0.3M & 90.3 & 92.2 & 92.8 & 93.3 & 93.6 & 93.9 & 94.0 & 94.1 & 94.2 & 93.2 \\
        & SCPNet \cite{scpnet} & - & 91.1 & 92.8 & \textbf{93.5} & 93.6 & 93.8 & 94.0 & 94.1 & 94.2 & 94.3 & 93.5 \\
        \cline{2-13}
        & Baseline & 20k & 90.5 & 92.2 & 92.8 & 93.0 & 93.3 & 93.8 & 93.9 & 94.0& 94.2 & 93.1 \\

        & Negative CoOp & 170k & 88.9 & 89.3 & 89.6 & 89.9 & 90.7 & 91.2 & 91.8 & 92.1& 92.4 & 90.8 \\

        & Positive CoOp & 170k & \textbf{91.4} & \textbf{92.8} & 93.4 & \textbf{93.6} & \textbf{93.8} & \textbf{94.0} & \textbf{94.2} & \textbf{94.2}& \textbf{94.3} & \textbf{93.6} \\

        \bottomrule
    \end{tabular}
    \caption{We compare the results of our three setups (Baseline, Positive CoOp, and Negative CoOp) with other SOTA approaches for MLR with partial annotations on COCO \cite{coco} and VOC 2007 \cite{pascal-voc}. The comparison is conducted with partial available labels (10\%-90\%) using the \textbf{ResNet101} architecture, following existing methods. The results demonstrate that the performance of the prompting-based approaches follows the order: Positive CoOp $>$ DualCoOp $\approx$ Baseline $>$ NegativeCoOp. * indicates method retrained using CLIP's weights.}
    \label{tab:mlr}
\end{table*}

\section{Experiments}
\label{sec:experiments}
\subsection{Dataset}
We evaluate the Baseline ($B$), PositiveCoOp and NegativeCoOp on standard MLR benchmark datasets: COCO \cite{coco} and VOC 2007 \cite{pascal-voc}. Our focus is on MLR with partial annotations. Consistent with previous approches \cite{SST, dualcoop, scpnet}, our experiments span a range of annotation availability, from 10\% to 90\% of the total labels.  Below, we outline the key details of these datasets and our approach for generating partial annotations:

 \noindent\textbf{MS-COCO 2014:}  COCO \cite{coco} is a large-scale popular multi-label recognition (MLR) dataset. The datasets consists of 80 classes belonging to various categories ranging from everyday objects like cars and people to animals and household items. The dataset contains 82,081 training images and 40,504 validation images. Consistent with existing MLR works, we use the training set for training and the validation set for inference.
\\
 \noindent \textbf{PASCAL VOC 2007:} VOC 2007 \cite{pascal-voc} is another widely used outdoor scene MLR dataset. It consists of 20 classes which overlap with the 80 classes from COCO dataset. It consists a total of 9,963 images belonging to those 20 classes. We follow the standard trainval set for training and use the test set for testing.

 To create training sets with partial labels from these datasets, we follow the methodology described in \cite{SST, dualcoop, scpnet}. Specifically, we randomly mask out a portion of the labels from the fully annotated training data, and use the remaining unmasked labels as the ground truth for training.
 
\subsection{Implementation Details}

For all our experiments, we use the original pretrained weights from  CLIP (Contrastive Language-Image Pre-Training) \cite{clip} as the VLM. Consistent with existing MLR literature \cite{dualcoop,scpnet,Cdul,dualcoop++,MLR-CoOp}, we use\textbf{ ResNet-101} as the visual encoder and the standard transformer for text encoder. Both visual and text encoders are frozen at all times. For a fair comparison, we use the same settings and hyperparameters as DualCoOp \cite{dualcoop}. We resize the images to $448$ for both datasets. And follow the augmentation methods Cutout \cite{cutout} and RandAugment \cite{randaug} to augment training images as described in \cite{dualcoop,scpnet,dualcoop++}. 
We train the context vectors $[V^+]$ and $ [V^-]$ of learnable prompts with stochastic gradient descent (SGD) using initial learning rate of 0.002. For PositiveCoOp and NegativeCoOp, we train the embeddings $\mathbf{r}_{j,-}$ and  $\mathbf{r}_{j,+}$ respectively,
using an initial learning rate of 1.0. For the baseline, we train the linear projector layer ($\Phi$) with SGD using initial learning rate of 0.01. All initial learning rates are reduced by cosine annealing for both datasets. Similar to DualCoOp \cite{dualcoop}, we train all setups for 50 epochs with batch size of 32. We set the loss hyperparameters in Eq. \ref{eq:instance_RASL} as $\gamma_- = 2$, $\gamma_+ = 1$ and $\delta$ = 0.05. We conduct all experiments on a single RTX A4000 GPU.

\subsection{Evaluation Metrics}
We evaluate our approach on the MLR datasets using the metric of mean average precision (mAP), as used by previous MLR approaches \cite{gcnmulti,dualcoop,scpnet,MLR-CoOp}. mAP is the mean of average precision (AP) values, where AP is computed as the area under the Precision-Recall curve for each class.

\subsection{Results}

We primarily compare the three setups (Baseline, PositiveCoOp and NegativeCoOp) with VLM-based MLR methods that operate in partial annotation settings \cite{dualcoop,scpnet} where the use of such an approach shows the greatest benefits. We do not compare with other MLR methods that are not tailored for partial label settings \cite{TaI-DPT,residual_attn_MLR,Pure_MLR_finetuning,Pure_MLR_finetuning,zhao2021m3tr,gao2021learning}. Our analysis mostly focuses on DualCoOp \cite{dualcoop}, because of its wide use, simplicity and focus on use of positive and negative prompts for MLR. 
We do not compare with DualCoOp++ \cite{dualcoop++}, its extension as (1) There is no publicly available code that reproduces their results and (2) They involve other components unrelated to negative prompting, making a specific ablation more difficult. However, we hypothesize that as they also use CLIP to learn a negative prompt, their performance could also benefit from a PositiveCoOp like setup.

Following previous work \cite{dualcoop, dualcoop++,scpnet, MLR-CoOp}, the evaluation is conducted across varying percentages of available labels from 10\% to 90\%.
Table \ref{tab:mlr}. presents a detailed comparison of the number of parameters required to train the network and the mean Average Precision (mAP) values achieved by our setups against the approaches that fit into the criteria described above, on the VOC 2007 \cite{pascal-voc} and COCO \cite{coco} datasets. 

The performance hierarchy of prompting-based approaches across the VOC2007 and COCO datasets follows: PositiveCoOp $>$ DualCoOp $\approx$ Baseline $>$ NegativeCoOp.
These results indicate:

\noindent(1) \textbf{Use of Negative prompting reduces performance:} PositiveCoOp achieves the best performance among the prompting-based methods. This implies that CLIP's guidance is beneficial for learning a prompt that can successfully detect the presence of classes. The poor performance of NegativeCoOp suggests that CLIP fails to guide the learning of negative prompts, making it ineffective in detecting the absence of classes.

(2) \textbf{Strong vision-only baseline performance:} Our vision-only baseline achieves comparable performance to DualCoOp while requiring approximately 15 times fewer training parameters and half the GPU hours across the two datasets. This suggests that baseline could be an optimal choice when the proportion of available labels is high and available compute is limited.


We report a comparison of parameters and GPU hours for training in Sec \edit{2} of the supplementary material.

\section{Analysis}
\label{Sec:Analysis}
%
In this section, we analyze why CLIP's guidance proves ineffective for negative prompt learning. We conduct a series of experiments for this, which follow the settings described in Sec \ref{sec:experiments} unless said otherwise.

%
\subsection{Presence of Negative Prompts in LAION-400M}
We hypothesize that negative prompts learnt using CLIP’s text encoder are not helpful for MLR because CLIP is not trained on images with such negative captions as images with such captions are rare on the internet from where CLIP’s training data is derived from. 
Simple examples of such positive and negative prompts to detect a dog would be 'A photo of a dog' and ' photo of a park not having a dog', and it is unlikely that the training set contained images of one object (e.g., a car) with captions describing the absence of another object (e.g., "not having a dog"). We conduct a series of experiments to test our hypothesis.

 To test our hypothesis, we analyze the LAION-400M dataset \cite{laion}, which comprises  400M image-text pairs derived using CLIP and has been used for training several Open Source VLMs such as OpenCLIP \cite{openclip}. Our analysis of the 413,871,335 texts revealed that only 1,961,669 texts (0.47\% of the total) contained negative words, confirming our hypothesis. Of these, 1,366,865 texts (0.33\% of the total) included a noun following the negative words, albeit not necessarily immediately after the negative word. This scarcity of negative text suggests that using CLIP's text encoder to learn negative prompts might not yield any performance benefits. Details on the list of negative words used and some examples of negative captions in LAION dataset are provided in Sec \edit{3.2} of the supplementary material. \\

\subsection{ Text Encoders Focus on Features Associated with Presence of a Class } 
If text encoder of VLM did provide useful guidance about features indicating absence of a class, text embeddings of handcrafted negative prompts of a class should be very dissimilar from positive prompts for the same classes. But due to the lack of negative captions in LAION400M, we hypothesize that such prompts will have embeddings that are strongly similar to positive prompts with the same noun because of the text encoders focus on nouns while ignoring the negations. 

To test this, we analyze similarity between CLIP embeddings of positive and negative prompts of 80 classes in the COCO dataset. Specifically,we use three prompts: positive prompt (P1): 'Photo of a \{classname\}', the corresponding negative prompt (N1): 'Not a photo of a \{classname\}' and another positive prompt (P2): 'Picture of a \{classname\}'. After passing them through CLIP’s text encoder, we compute the cosine similarity embeddings of P1 and P2, and between those of P1 and N1. These are averaged across all classes. The results in Table \ref{tab:cosine_similarity} show that P1 and N1 are almost as similar as P1-P2 implying that both positive and negative prompts are projected closely in the feature space. This validates our hypothesis, suggesting that the text encoder is unable identify embedding features associated with the absence of the class and instead focuses on embedding features associated with presence of the noun (class). Fig.\ref{fig:CLIP's similarity}  visualizes the similarity of CLIP features from different regions of an image to embeddings of such positive and negative prompts which also show that both prompts are similar to the regions having the object, supporting our hypothesis. Additional results are provided in Fig \edit{1}. of the supplementary material. We also conduct a more extensive experiment with multiple commonly used prompt formats that yields similar results. More details are reported in Sec \edit{3.1} of the supplementary material.

\begin{table}[t]
\centering
\begin{tabular}{p{0.3\linewidth}|p{0.3\linewidth}|p{0.3\linewidth}}
\hline
Cosine Similarity \newline (80 cls-1 prompt) & P1:'photo of a\{\}' \newline N1:'Not a photo of a \{\}' & P1:'photo of a\{\}' \newline P2:'picture of a \{\}' \\
\hline
Mean $\pm$ Std & 0.58 $\pm$ 0.06 & 0.53 $\pm$ 0.04\\
(Min,Max) & (0.37, 0.69) & (0.51, 0.67) \\
\hline
\hline
Cosine Similarity \newline(80cls-85prompt) & P1-N1 Pairs  & P1-P2 Pairs \\
\hline
Mean $\pm$ Std & 0.56 $\pm$ 0.06 & 0.61 $\pm$ 0.01\\

(Min, Max) & (0.37, 0.67) & (0.55, 0.63) \\
\hline
\end{tabular}
\caption{ \textbf{Cosine similarity between prompt features.} We compare the average similarity between pairs of positive features and pairs of positive and negative on 80 classes of COCO dataset for two scenarios a) When we use a only one prompt and b) Using 85 default prompt templates for ImageNet. We observe that the similarity score between positive-positive prompt is close to positive-negative, implying that CLIP projects positive and negative prompts very closely in the feature space.}

\label{tab:cosine_similarity}
\end{table}

\section{Conclusions}
In this paper, we examined the role of prompt learning in VLM-based multi-label recognition (MLR) with partial annotations. We specifically estimate the contribution of positive and negative prompts to MLR separately by using our ablated MLR setups: PositiveCoOp and NegativeCoOp, where one prompt is learned under VLM guidance while the other is represented by a learned embedding in the shared feature space. Our results show that learning only positive prompts while using learned negative embeddings (PositiveCoOp) consistently outperforms dual prompt learning approaches, indicating that learning negative prompts for MLR using VLM guidance degrades performance. Our analysis of the LAION-400M points to the lack of negative prompts in the dataset as the likely reason for this. Additionally, we found that in settings with a low proportion of missing labels, a vision-features-only baseline shows significantly strong performance while being much more efficient in terms of computation time (GPU hours) and parameters.



{\small
\bibliographystyle{ieee_fullname}
\bibliography{PaperForReview}

\begin{thebibliography}{10}\itemsep=-1pt

\bibitem{Cdul}
Rabab Abdelfattah, Qing Guo, Xiaoguang Li, Xiaofeng Wang, and Song Wang.
\newblock Cdul: Clip-driven unsupervised learning for multi-label image classification.
\newblock In {\em Proceedings of the IEEE/CVF International Conference on Computer Vision}, pages 1348--1357, 2023.

\bibitem{nutrient3}
Marios~M Anthimopoulos, Lauro Gianola, Luca Scarnato, Peter Diem, and Stavroula~G Mougiakakou.
\newblock A food recognition system for diabetic patients based on an optimized bag-of-features model.
\newblock {\em IEEE journal of biomedical and health informatics}, 18(4):1261--1271, 2014.

\bibitem{ret}
Shubhang Bhatnagar and Narendra Ahuja.
\newblock Piecewise-linear manifolds for deep metric learning.
\newblock In {\em Conference on Parsimony and Learning}, pages 269--281. PMLR, 2024.

\bibitem{ecomm_product}
Wei-Cheng Chang, Daniel Jiang, Hsiang-Fu Yu, Choon~Hui Teo, Jiong Zhang, Kai Zhong, Kedarnath Kolluri, Qie Hu, Nikhil Shandilya, Vyacheslav Ievgrafov, et~al.
\newblock Extreme multi-label learning for semantic matching in product search.
\newblock In {\em Proceedings of the 27th ACM SIGKDD conference on knowledge discovery \& data mining}, pages 2643--2651, 2021.

\bibitem{kggr}
Tianshui Chen, Liang Lin, Riquan Chen, Xiaolu Hui, and Hefeng Wu.
\newblock Knowledge-guided multi-label few-shot learning for general image recognition.
\newblock {\em IEEE Transactions on Pattern Analysis and Machine Intelligence}, 44(3):1371--1384, 2020.

\bibitem{SST}
Tianshui Chen, Tao Pu, Hefeng Wu, Yuan Xie, and Liang Lin.
\newblock Structured semantic transfer for multi-label recognition with partial labels.
\newblock In {\em Proceedings of the AAAI conference on artificial intelligence}, volume~36, pages 339--346, 2022.

\bibitem{SSGRL}
Tianshui Chen, Muxin Xu, Xiaolu Hui, Hefeng Wu, and Liang Lin.
\newblock Learning semantic-specific graph representation for multi-label image recognition.
\newblock In {\em Proceedings of the IEEE/CVF international conference on computer vision}, pages 522--531, 2019.

\bibitem{gcnmulti}
Zhao-Min Chen, Xiu-Shen Wei, Peng Wang, and Yanwen Guo.
\newblock Multi-label image recognition with graph convolutional networks.
\newblock In {\em Proceedings of the IEEE/CVF conference on computer vision and pattern recognition}, pages 5177--5186, 2019.

\bibitem{cole2021multi}
Elijah Cole, Oisin Mac~Aodha, Titouan Lorieul, Pietro Perona, Dan Morris, and Nebojsa Jojic.
\newblock Multi-label learning from single positive labels.
\newblock In {\em Proceedings of the IEEE/CVF Conference on Computer Vision and Pattern Recognition}, pages 933--942, 2021.

\bibitem{randaug}
Ekin~D Cubuk, Barret Zoph, Jonathon Shlens, and Quoc~V Le.
\newblock Randaugment: Practical automated data augmentation with a reduced search space.
\newblock In {\em Proceedings of the IEEE/CVF conference on computer vision and pattern recognition workshops}, pages 702--703, 2020.

\bibitem{deng2009imagenet}
Jia Deng, Wei Dong, Richard Socher, Li-Jia Li, Kai Li, and Li Fei-Fei.
\newblock Imagenet: A large-scale hierarchical image database.
\newblock In {\em 2009 IEEE conference on computer vision and pattern recognition}, pages 248--255. Ieee, 2009.

\bibitem{cutout}
Terrance DeVries and Graham~W Taylor.
\newblock Improved regularization of convolutional neural networks with cutout.
\newblock {\em arXiv preprint arXiv:1708.04552}, 2017.

\bibitem{scpnet}
Zixuan Ding, Ao Wang, Hui Chen, Qiang Zhang, Pengzhang Liu, Yongjun Bao, Weipeng Yan, and Jungong Han.
\newblock Exploring structured semantic prior for multi label recognition with incomplete labels.
\newblock In {\em Proceedings of the IEEE/CVF Conference on Computer Vision and Pattern Recognition}, pages 3398--3407, 2023.

\bibitem{partial-label-learning3}
Thibaut Durand, Nazanin Mehrasa, and Greg Mori.
\newblock Learning a deep convnet for multi-label classification with partial labels.
\newblock In {\em Proceedings of the IEEE/CVF conference on computer vision and pattern recognition}, pages 647--657, 2019.

\bibitem{pascal-voc}
Mark Everingham, Luc Van~Gool, Christopher~KI Williams, John Winn, and Andrew Zisserman.
\newblock The pascal visual object classes (voc) challenge.
\newblock {\em International journal of computer vision}, 88:303--338, 2010.

\bibitem{gao2021learning}
Bin-Bin Gao and Hong-Yu Zhou.
\newblock Learning to discover multi-class attentional regions for multi-label image recognition.
\newblock {\em IEEE Transactions on Image Processing}, 30:5920--5932, 2021.

\bibitem{adapter1}
Peng Gao, Shijie Geng, Renrui Zhang, Teli Ma, Rongyao Fang, Yongfeng Zhang, Hongsheng Li, and Yu Qiao.
\newblock Clip-adapter: Better vision-language models with feature adapters.
\newblock {\em International Journal of Computer Vision}, 132(2):581--595, 2024.

\bibitem{TaI-DPT}
Zixian Guo, Bowen Dong, Zhilong Ji, Jinfeng Bai, Yiwen Guo, and Wangmeng Zuo.
\newblock Texts as images in prompt tuning for multi-label image recognition.
\newblock In {\em Proceedings of the IEEE/CVF Conference on Computer Vision and Pattern Recognition}, pages 2808--2817, 2023.

\bibitem{dualcoop++}
Ping Hu, Ximeng Sun, Stan Sclaroff, and Kate Saenko.
\newblock Dualcoop++: Fast and effective adaptation to multi-label recognition with limited annotations.
\newblock {\em IEEE Transactions on Pattern Analysis and Machine Intelligence}, 2023.

\bibitem{huang2024radiology}
Haoxu Huang, Samyak Rawlekar, Sumit Chopra, and Cem~M Deniz.
\newblock Radiology reports improve visual representations learned from radiographs.
\newblock In {\em Medical Imaging with Deep Learning}, pages 1385--1405. PMLR, 2024.

\bibitem{openclip}
Gabriel Ilharco, Mitchell Wortsman, Ross Wightman, Cade Gordon, Nicholas Carlini, Rohan Taori, Achal Dave, Vaishaal Shankar, Hongseok Namkoong, John Miller, Hannaneh Hajishirzi, Ali Farhadi, and Ludwig Schmidt.
\newblock Openclip, July 2021.
\newblock If you use this software, please cite it as below.

\bibitem{partial-label-learning2}
Armand Joulin, Laurens Van Der~Maaten, Allan Jabri, and Nicolas Vasilache.
\newblock Learning visual features from large weakly supervised data.
\newblock In {\em Computer Vision--ECCV 2016: 14th European Conference, Amsterdam, The Netherlands, October 11--14, 2016, Proceedings, Part VII 14}, pages 67--84. Springer, 2016.

\bibitem{ret2}
Shyamgopal Karthik, Karsten Roth, Massimiliano Mancini, and Zeynep Akata.
\newblock Vision-by-language for training-free compositional image retrieval.
\newblock {\em arXiv preprint arXiv:2310.09291}, 2023.

\bibitem{focal}
Tsung-Yi Lin, Priya Goyal, Ross Girshick, Kaiming He, and Piotr Doll{\'a}r.
\newblock Focal loss for dense object detection.
\newblock In {\em Proceedings of the IEEE international conference on computer vision}, pages 2980--2988, 2017.

\bibitem{coco}
Tsung-Yi Lin, Michael Maire, Serge Belongie, James Hays, Pietro Perona, Deva Ramanan, Piotr Doll{\'a}r, and C~Lawrence Zitnick.
\newblock Microsoft coco: Common objects in context.
\newblock In {\em Computer Vision--ECCV 2014: 13th European Conference, Zurich, Switzerland, September 6-12, 2014, Proceedings, Part V 13}, pages 740--755. Springer, 2014.

\bibitem{rnn1}
Feng Liu, Tao Xiang, Timothy~M Hospedales, Wankou Yang, and Changyin Sun.
\newblock Semantic regularisation for recurrent image annotation.
\newblock In {\em Proceedings of the IEEE Conference on Computer Vision and Pattern Recognition}, pages 2872--2880, 2017.

\bibitem{liu2015optimality}
Weiwei Liu and Ivor Tsang.
\newblock On the optimality of classifier chain for multi-label classification.
\newblock {\em Advances in Neural Information Processing Systems}, 28, 2015.

\bibitem{partial-label-learning1}
Dhruv Mahajan, Ross Girshick, Vignesh Ramanathan, Kaiming He, Manohar Paluri, Yixuan Li, Ashwin Bharambe, and Laurens Van Der~Maaten.
\newblock Exploring the limits of weakly supervised pretraining.
\newblock In {\em Proceedings of the European conference on computer vision (ECCV)}, pages 181--196, 2018.

\bibitem{nutrient2}
Austin Meyers, Nick Johnston, Vivek Rathod, Anoop Korattikara, Alex Gorban, Nathan Silberman, Sergio Guadarrama, George Papandreou, Jonathan Huang, and Kevin~P Murphy.
\newblock Im2calories: towards an automated mobile vision food diary.
\newblock In {\em Proceedings of the IEEE international conference on computer vision}, pages 1233--1241, 2015.

\bibitem{misra2016seeing}
Ishan Misra, C Lawrence~Zitnick, Margaret Mitchell, and Ross Girshick.
\newblock Seeing through the human reporting bias: Visual classifiers from noisy human-centric labels.
\newblock In {\em Proceedings of the IEEE conference on computer vision and pattern recognition}, pages 2930--2939, 2016.

\bibitem{SARB}
Tao Pu, Tianshui Chen, Hefeng Wu, and Liang Lin.
\newblock Semantic-aware representation blending for multi-label image recognition with partial labels.
\newblock In {\em Proceedings of the AAAI conference on artificial intelligence}, volume~36, pages 2091--2098, 2022.

\bibitem{clip}
Alec Radford, Jong~Wook Kim, Chris Hallacy, Aditya Ramesh, Gabriel Goh, Sandhini Agarwal, Girish Sastry, Amanda Askell, Pamela Mishkin, Jack Clark, et~al.
\newblock Learning transferable visual models from natural language supervision.
\newblock In {\em International conference on machine learning}, pages 8748--8763. PMLR, 2021.

\bibitem{MLR-CoOp}
Samyak Rawlekar, Shubhang Bhatnagar, Vishnuvardhan~Pogunulu Srinivasulu, and Narendra Ahuja.
\newblock Improving multi-label recognition using class co-occurrence probabilities.
\newblock {\em arXiv preprint arXiv:2404.16193}, 2024.

\bibitem{asl}
Tal Ridnik, Emanuel Ben-Baruch, Nadav Zamir, Asaf Noy, Itamar Friedman, Matan Protter, and Lihi Zelnik-Manor.
\newblock Asymmetric loss for multi-label classification.
\newblock In {\em Proceedings of the IEEE/CVF International Conference on Computer Vision}, pages 82--91, 2021.

\bibitem{boostexter}
Robert~E Schapire and Yoram Singer.
\newblock Boostexter: A boosting-based system for text categorization.
\newblock {\em Machine learning}, 39:135--168, 2000.

\bibitem{laion}
Christoph Schuhmann, Richard Vencu, Romain Beaumont, Robert Kaczmarczyk, Clayton Mullis, Aarush Katta, Theo Coombes, Jenia Jitsev, and Aran Komatsuzaki.
\newblock Laion-400m: Open dataset of clip-filtered 400 million image-text pairs.
\newblock {\em arXiv preprint arXiv:2111.02114}, 2021.

\bibitem{dualcoop}
Ximeng Sun, Ping Hu, and Kate Saenko.
\newblock Dualcoop: Fast adaptation to multi-label recognition with limited annotations.
\newblock {\em Advances in Neural Information Processing Systems}, 35:30569--30582, 2022.

\bibitem{tsoumakas2006multi}
G Tsoumakas and I Katakis.
\newblock Multi-label classification: An overview international journal of data warehousing and mining.
\newblock {\em The label powerset algorithm is called PT3}, 3(3), 2006.

\bibitem{rnn2}
Jiang Wang, Yi Yang, Junhua Mao, Zhiheng Huang, Chang Huang, and Wei Xu.
\newblock Cnn-rnn: A unified framework for multi-label image classification.
\newblock In {\em Proceedings of the IEEE conference on computer vision and pattern recognition}, pages 2285--2294, 2016.

\bibitem{Pure_MLR_finetuning}
Ya Wang, Dongliang He, Fu Li, Xiang Long, Zhichao Zhou, Jinwen Ma, and Shilei Wen.
\newblock Multi-label classification with label graph superimposing.
\newblock In {\em Proceedings of the AAAI Conference on Artificial Intelligence}, volume~34, pages 12265--12272, 2020.

\bibitem{rare}
Jeremy~M Wolfe, Todd~S Horowitz, and Naomi~M Kenner.
\newblock Rare items often missed in visual searches.
\newblock {\em Nature}, 435(7041):439--440, 2005.

\bibitem{adapter2}
Mitchell Wortsman, Gabriel Ilharco, Jong~Wook Kim, Mike Li, Simon Kornblith, Rebecca Roelofs, Raphael~Gontijo Lopes, Hannaneh Hajishirzi, Ali Farhadi, Hongseok Namkoong, et~al.
\newblock Robust fine-tuning of zero-shot models.
\newblock In {\em Proceedings of the IEEE/CVF conference on computer vision and pattern recognition}, pages 7959--7971, 2022.

\bibitem{seg2}
Mengde Xu, Zheng Zhang, Fangyun Wei, Yutong Lin, Yue Cao, Han Hu, and Xiang Bai.
\newblock A simple baseline for open-vocabulary semantic segmentation with pre-trained vision-language model.
\newblock In {\em European Conference on Computer Vision}, pages 736--753. Springer, 2022.

\bibitem{adapter3}
Yuan Yao, Ao Zhang, Zhengyan Zhang, Zhiyuan Liu, Tat-Seng Chua, and Maosong Sun.
\newblock Cpt: Colorful prompt tuning for pre-trained vision-language models.
\newblock {\em AI Open}, 5:30--38, 2024.

\bibitem{rnn4}
Vacit~Oguz Yazici, Abel Gonzalez-Garcia, Arnau Ramisa, Bartlomiej Twardowski, and Joost van~de Weijer.
\newblock Orderless recurrent models for multi-label classification.
\newblock In {\em Proceedings of the IEEE/CVF Conference on Computer Vision and Pattern Recognition}, pages 13440--13449, 2020.

\bibitem{adapter4}
Renrui Zhang, Wei Zhang, Rongyao Fang, Peng Gao, Kunchang Li, Jifeng Dai, Yu Qiao, and Hongsheng Li.
\newblock Tip-adapter: Training-free adaption of clip for few-shot classification.
\newblock In {\em European conference on computer vision}, pages 493--510. Springer, 2022.

\bibitem{zhao2021m3tr}
Jiawei Zhao, Yifan Zhao, and Jia Li.
\newblock M3tr: Multi-modal multi-label recognition with transformer.
\newblock In {\em Proceedings of the 29th ACM international conference on multimedia}, pages 469--477, 2021.

\bibitem{residual_attn_MLR}
Ke Zhu and Jianxin Wu.
\newblock Residual attention: A simple but effective method for multi-label recognition.
\newblock In {\em Proceedings of the IEEE/CVF international conference on computer vision}, pages 184--193, 2021.

\end{thebibliography}
}
\clearpage

\clearpage
\twocolumn[
\begin{center}
\textbf{\Huge Supplementary Material}
\vspace{0.3in}
\end{center}
]

\setcounter{section}{0}
\setcounter{equation}{0}
\setcounter{figure}{0}


\section{Additional Visual Results}
\subsection{Prompt-Image Embedding Similarity Map}

\begin{figure*}
    \centering
    \includegraphics[width=0.75\linewidth]{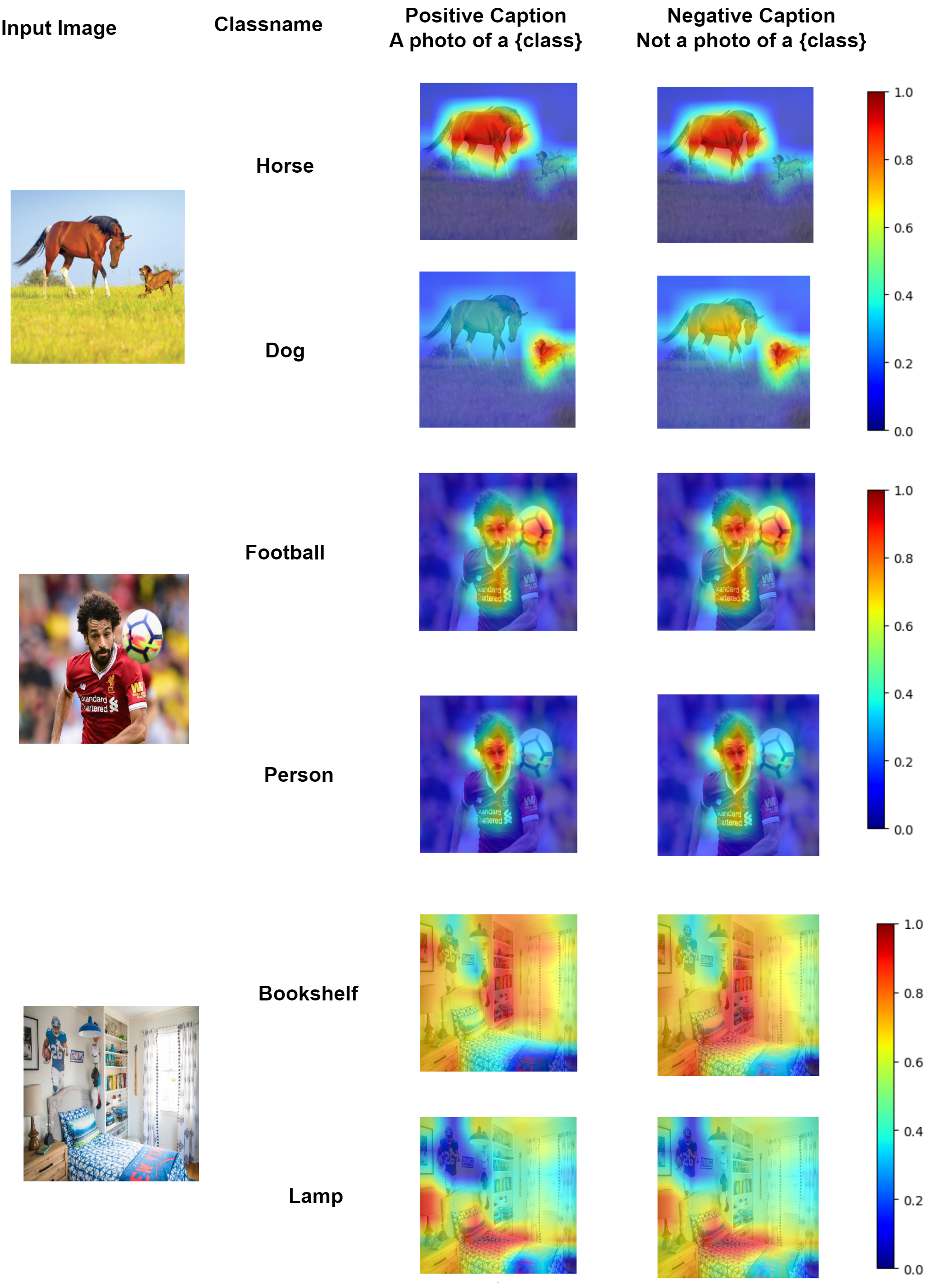}
    \caption{\textbf{Visualization of Similarity Maps.}
    We compare similarity maps obtained using cosine similarity between image features and positive prompt features versus image features and negative prompt features for each class. The activation of similar regions in both maps questions the effectiveness of CLIP's guidance for learning a negative prompt.}
    \label{fig:prompt_vis}
\end{figure*}

In Fig.\ref{fig:prompt_vis}, we provide additional visualizations comparing the similarity maps generated by the calculating the cosine similarity between image features and positive prompt features and image features and  negative prompts features for each class. Consistent with the main paper, for each class, both the positive and negative prompt similarity maps activate the same regions that are associated with the presence of the class.

\subsection{LAION-400M Dataset Visualization}
\begin{figure*}
    \centering
    \includegraphics[width=1\linewidth]{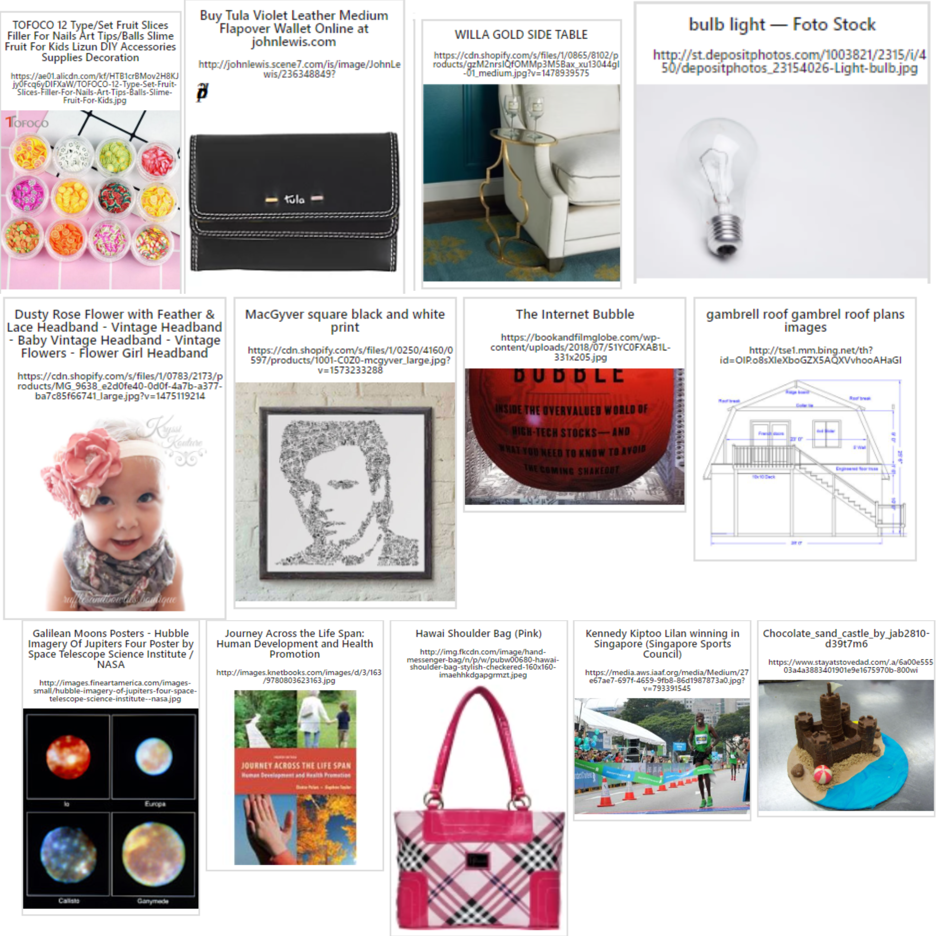}
    \caption{Image Text pairs from LAION400M dataset. The descriptions of the images mainly focus on the objects(classes) present in the image, and do not describe the absence of objects (classes).}
    \label{fig:LAION EXAMPLE}
\end{figure*}

In Fig.\ref{fig:LAION EXAMPLE}, we provide examples of image-text pairs from the LAION-400M dataset\cite{laion}. We observe that text descriptions mainly focus on the presence of objects(class) and do not describe the absence of objects(class) not present in the image. This explains why CLIP struggles to effectively guide the learning of a negative prompt.




%
\section{Computation Comparison}

In this section, we present a quantitative comparison of the number of parameters and GPU hours required to train DualCoOp \cite{dualcoop}, SCPNet \cite{scpnet}, and the setups proposed in our paper: Baseline, PositiveCoOp, and NegativeCoOp. The experiments were conducted on a single NVIDIA RTX A4000 GPU using the COCO \cite{coco} and VOC2007 \cite{pascal-voc} datasets. We observe that, compared to DualCoOp (fewest parameters and training hours among existing methods), the Baseline requires approximately 15 times fewer parameters and half the GPU hours on both datasets. This is primarily because the Baseline does not backpropagate through the text encoder as DualCoOp does, significantly reducing the computation requirement. PositiveCoOp outperforms DualCoOp while requiring approximately 0.6 times fewer parameters. The results are detailed in Table \ref{tab:compute_time}.

\begin{table}[H]
\centering
\begin{tabular}{l|l|c|c}
\hline
\textbf{Dataset} & \textbf{Method} & \textbf{\#Params} & \textbf{GPU Hours} \\ \hline
\multirow{3}{*}{VOC2007} & DualCoOp \cite{dualcoop} & 0.3M & 3.55 \\ 
 & SCPNet \cite{scpnet}& - & 3 \\
 & Baseline & 20k & 1.5 \\
 & NegativeCoOp & 0.17M & 3 \\ 
 & PositiveCoOp & 0.17M & 3 \\ \hline
\multirow{3}{*}{COCO} & DualCoOp \cite{dualcoop} & 1.3M & 16 \\ 
 & SCPNet \cite{scpnet} & 3.4M & 26 \\ 
 & Baseline & 80k& 7.97 \\ 
 & NegativeCoOp & 0.73M& 16 \\
 & PositiveCoOp & 0.73M& 16 \\ \hline
\end{tabular}
\caption{\textbf{Computation Comparison}: We compare the training parameters and GPU hours for Baseline, PositiveCoOp, and NegativeCoOp with existing VLM-based MLR methods in partial annotation settings. Baseline uses significantly fewer parameters and GPU hours than all other setups, while PositiveCoOp and NegativeCoOp require about half the parameters compared to DualCoOp. }
\label{tab:compute_time}
\end{table}

\section{Details of Experiments in Sec. 5 of Paper}

\subsection{Text Encoders Focus on Features Associated
with Presence of a Class}

In Table.\ref{85_prompts}, we list all 85 prompts that were used to generate the quantitative results in Sec \textbf{5.2}

\begin{table*}[h!]
\centering
\begin{tabular}{|p{0.3\linewidth}|p{0.3\linewidth}|p{0.3\linewidth}|}
\hline
a bad photo of a \{\}. & a photo of many \{\}. & a sculpture of a \{\}. \\
\hline
a photo of the hard to see \{\}. & a low resolution photo of the \{\}. & a rendering of a \{\}. \\
\hline
graffiti of a \{\}. & a bad photo of the \{\}. & a cropped photo of the \{\}. \\
\hline
a tattoo of a \{\}. & the embroidered \{\}. & a photo of a hard to see \{\}. \\
\hline
a bright photo of a \{\}. & a photo of a clean \{\}. & a photo of a dirty \{\}. \\
\hline
a dark photo of the \{\}. & a drawing of a \{\}. & a photo of my \{\}. \\
\hline
the plastic \{\}. & a photo of the cool \{\}. & a close-up photo of a \{\}. \\
\hline
a black and white photo of the \{\}. & a painting of the \{\}. & a painting of a \{\}. \\
\hline
a pixelated photo of the \{\}. & a sculpture of the \{\}. & a bright photo of the \{\}. \\
\hline
a cropped photo of a \{\}. & a plastic \{\}. & a photo of the dirty \{\}. \\
\hline
a jpeg corrupted photo of a \{\}. & a blurry photo of the \{\}. & a photo of the \{\}. \\
\hline
a good photo of the \{\}. & a rendering of the \{\}. & a \{\} in a video game. \\
\hline
a photo of one \{\}. & a doodle of a \{\}. & a close-up photo of the \{\}. \\
\hline
a photo of a \{\}. & the origami \{\}. & the \{\} in a video game. \\
\hline
a sketch of a \{\}. & a doodle of the \{\}. & a origami \{\}. \\
\hline
a low resolution photo of a \{\}. & the toy \{\}. & a rendition of the \{\}. \\
\hline
a photo of the clean \{\}. & a photo of a large \{\}. & a rendition of a \{\}. \\
\hline
a photo of a nice \{\}. & a photo of a weird \{\}. & a blurry photo of a \{\}. \\
\hline
a cartoon \{\}. & art of a \{\}. & a sketch of the \{\}. \\
\hline
a embroidered \{\}. & a pixelated photo of a \{\}. & itap of the \{\}. \\
\hline
a jpeg corrupted photo of the \{\}. & a good photo of a \{\}. & a plushie \{\}. \\
\hline
a photo of the nice \{\}. & a photo of the small \{\}. & a photo of the weird \{\}. \\
\hline
the cartoon \{\}. & art of the \{\}. & a drawing of the \{\}. \\
\hline
a photo of the large \{\}. & a black and white photo of a \{\}. & the plushie \{\}. \\
\hline
a dark photo of a \{\}. & itap of a \{\}. & graffiti of the \{\}. \\
\hline
a toy \{\}. & itap of my \{\}. & a photo of a cool \{\}. \\
\hline
a photo of a small \{\}. & a tattoo of the \{\}. & there is a \{\} in the scene. \\
\hline
there is the \{\} in the scene. & this is a \{\} in the scene. & this is the \{\} in the scene. \\
\hline
this is one \{\} in the scene. & & \\
\hline
\end{tabular}
\caption{\textbf{Prompt Templates}: Provides the complete list of prompts (85 default prompts from ImageNet \cite{deng2009imagenet}) that were used to find the similarity between prompt features.}
\label{85_prompts}
\end{table*}

\subsection{Presence of Negative Prompts in LAION-400M}

In this section, we provide the details for Sec. \textcolor{red}{5.1} of the main paper, where we show that less than 0.5\% of the LAION400M dataset has negative words, and even fewer have negative words followed by a noun (not necessarily immediately). The list of negative words we used for the task is as follows:

\begin{center}
\begin{minipage}{0.9\columnwidth}
\begin{verbatim}
Negative Words = {
    'not', 'no', 'never', 
    'none', 'nothing', 'nobody'
    'nowhere', 'neither', 'nor',
    "can't", "cannot", "won't",
    "don't", "doesn't", "didn't", 
    "isn't", "aren't", "wasn't",
    "weren't", "hasn't", "haven't",
    "hadn't", "shouldn't", "wouldn't",
    "couldn't", "mustn't"
}
\end{verbatim}
\end{minipage}
\end{center}

Below, we provide some examples of text containing negative words and the phrases that include the negative word followed by a noun, referred to as 'Phrase'.
\begin{enumerate}
    \item Post-it note saying I will not compare myself to a stranger on Instagram \\
    \textbf{Phrase}: not compare myself to a stranger
    \item I'm not getting any younger Magnet \\
    \textbf{Phrase}: not getting any younger Magnet
    \item Lionel I was told this cake was not cut. It was painted with butter Cream Icing right on the cake. No transfer.
    \textbf{Phrase}: No transfer

    \item Dennis Pitta was not targeted often, but made his presence felt against Denver. \\
\textbf{Phrase}: not targeted often, but made his presence
    \item Introduction Video recommendation in YouTube Related to the current video topic and user profile but not visualized \\
    \textbf{Phrase:} -
        \item Arguing, Children, and Facts: AMERICAN SNIPER  Follow  NEMA  ACTS @cinfacts  for more content  CHRIS KYLE IS DEPICTED IN THE MOVIE AS SHOOTING THE  CHILD WITH THE GRENADE AND THEN THE MOTHER WHEN SHE  PICKED IT UP TO THROW IT. BUT IN THE REAL INCIDENT CHRIS  KYLE NEVER SHOT ANY CHILDREN. SHE WAS THE ONLY ONE  SHOT WHEN SHE ATTEMPTED TO ATTACK U.S. FORCES. People will argue, ""It\'s just a kid!"" Well, I\'d agree with you but if I\'m a soldier and some kid runs at me with a live explosive, I\'m not going to hesitate. Your thoughts? - Follow @cinfacts for more facts. \\
    \textbf{Phrase:} NEVER SHOT
    \item Yes i do.......but not with you ! aprons \\
    \textbf{Phrase}: not with you ! aprons
    \item Scary creatures: Brown's monster-strewn Hollywood Hills mansion not surprisingly distressed his wealthy neighbours \\
    \textbf{Phrase}: not surprisingly distressed his wealthy neighbours

    \item An elderly monk meditates before the historic Shwedagon Paya, Yangon, Unuon of Myanmar (Burma), Nov. 26, 2009. The Shwedagon pagoda's central hti, an umbrella spire atop the giant zedi structure, sports a 76-karat diamond that casts red, green white beams to specific spots on the terrace as the sun rises or sets. The massive complex sits atop a 190-foot hill accessed by four stair-stepped walkways guarded by 30-foot-tall mythical half-lion half-dragon creatures called chinthe. The central 98-foot-tall zedi is surrounded by an incredible assortment of other smaller zedi, statues and temples...EDS: Not for syndication nor redistribution. Web slide show only. Please do not strip metadata for Web use. \\
    \textbf{Phrase}: nor redistribution

    \item KEEP CALM AND REMMEMBER You are not TUNA - Personalised iPhone 6 / 6S Case: Full Wrap White \\
    \textbf{Phrase}: not TUNA
    \item  Problems with the MMSE Mini-Mental State Exam – no psychometrics –Folstein et al., 1975 (antique) Considerable noise Several items do not provide adequate information Poor range for measuring change –Large standard error of measurement Poor power for assessing medication benefit Inadequate screening tool Better, shorter tests are available Now, copyright is being enforced (not free!!) \\
    \textbf{Phrase} - no psychometrics
    
\end{enumerate}



\end{document}